\documentclass[10pt,twocolumn]{article}
\usepackage[twoside,letterpaper,includeheadfoot,%
        layoutsize={8.125in,10.875in},%
                layouthoffset=0.1875in,%
                layoutvoffset=0.0625in,%
                left=43pt,%
                right=43pt,%
                top=43pt,% 10pt provided by headsep
                bottom=43pt,%
                headheight=0pt,% No Header
                headsep=10pt,%
                footskip=25pt]{geometry}

\usepackage{graphicx,amsmath,amsfonts,amssymb}
\usepackage{fancyhdr}
\usepackage[footnotesize]{caption}
\usepackage{outlines}
\usepackage{apacite}
\usepackage{multirow}
\usepackage{tabularx}
\usepackage{booktabs}
\usepackage{array}

\usepackage{soul} % to use \hl
\usepackage{color}
\usepackage{xcolor,colortbl}
\definecolor{orange}{HTML}{FCBF64}
\definecolor{skin}{HTML}{E79E97}

\usepackage{natbib}
\usepackage[explicit]{titlesec}
\titleformat{\section}
  {\large\bfseries}
  {\thesection.}
  {0.5em}
  {#1}
  []
\titleformat{name=\section,numberless}
  {\large\bfseries}
  {}
  {0em}
  {#1}
  []  
\titleformat{\subsection}[runin]
  {\bfseries}
  {\thesubsection.}
  {0.5em}
  {#1. }
  []
\titleformat{\subsubsection}[runin]
  {\small\bfseries\itshape}
  {\thesubsubsection.}
  {0.5em}
  {#1. }
  []    
\titleformat{\paragraph}[runin]
  {\small\bfseries}
  {}
  {0em}
  {#1} 
\titlespacing*{\section}{0pc}{3ex \@plus4pt \@minus3pt}{5pt}
\titlespacing*{\subsection}{0pc}{2.5ex \@plus3pt \@minus2pt}{2pt}
\titlespacing*{\subsubsection}{0pc}{2ex \@plus2.5pt \@minus1.5pt}{2pt}
\titlespacing*{\paragraph}{0pc}{1.5ex \@plus2pt \@minus1pt}{12pt}

\title{\vspace{0.4cm} \LARGE \textbf{The Omniglot challenge: a 3-year progress report} }
\date{}
\author{ \large \textbf{Brenden M. Lake}$^1$, \textbf{Ruslan Salakhutdinov}$^2$, and \textbf{Joshua B. Tenenbaum}$^{3}$\\
\normalsize $^1$Department of Psychology and Center for Data Science, New York University \\
\normalsize $^2$Machine Learning Department, Carnegie Mellon University \\
\normalsize $^3$Department of Brain and Cognitive Sciences and Center for Brains Minds and Machines, MIT \\
}

\fancypagestyle{alim}{\fancyhf{}\fancyhead[L]{In press at \emph{Current Opinion in Behavioral Sciences}.}}

\begin{document}
\maketitle
\thispagestyle{alim}

\begin{abstract}
Three years ago, we released the Omniglot dataset for one-shot learning, along with five challenge tasks and a computational model that addresses these tasks. The model was not meant to be the final word on Omniglot; we hoped that the community would build on our work and develop new approaches. In the time since, we have been pleased to see wide adoption of the dataset. There has been notable progress on one-shot classification, but researchers have adopted new splits and procedures that make the task easier. There has been less progress on the other four tasks. We conclude that recent approaches are still far from human-like concept learning on Omniglot, a challenge that requires performing many tasks with a single model.
\end{abstract}

\section*{Introduction}
Three years ago, we released the Omniglot dataset of handwritten characters from 50 different alphabets \citep{LakeScience2015}. The dataset was developed to study how humans and machines perform one-shot learning -- the ability to learn a new concept from just a single example. The domain of handwritten characters provides a large set of novel, high-dimensional concepts that people learn and use in the real world. Characters are far more complex than the low-dimensional artificial stimuli used in classic psychological studies of concept learning \citep{Bruner1956,Shepard1961}, and they are still simple and tractable enough to hope that machines, in the near future, will see most of the structure in the images that people do. For these reasons, Omniglot is an ideal testbed for studying human and machine learning, and it was released as a challenge to the cognitive science, machine learning, and artificial intelligence (AI) communities.

\begin{figure*}[t]
\centering
\includegraphics[width=5in]{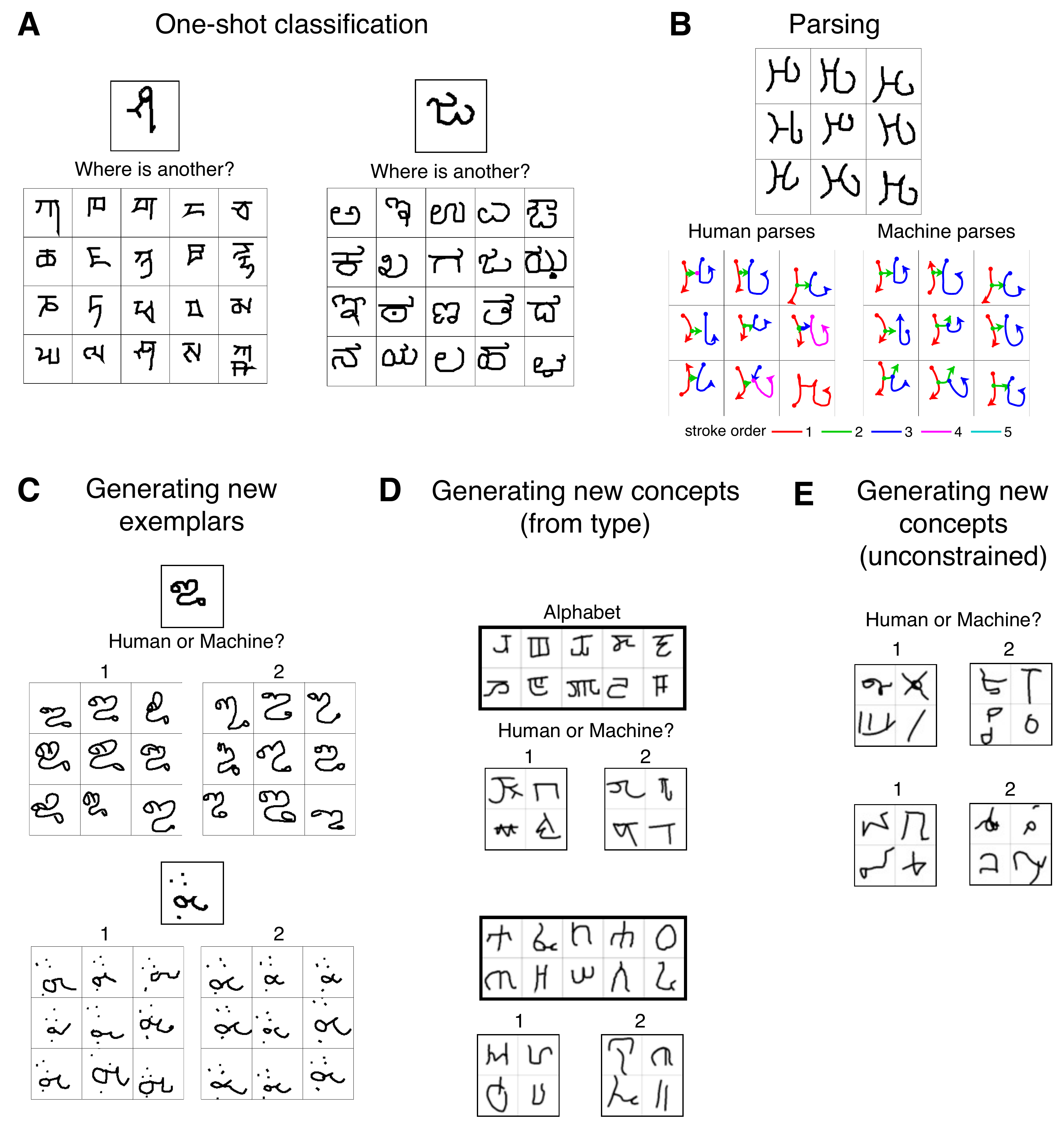}
\caption{The Omniglot challenge of performing five concept learning tasks at a human level. A) Two trials of one-shot classification, where a single image of a new character is presented (top) and the goal is to select another example of that character amongst other characters from the same alphabet (in the grid below). In panels B)-E), human participants and Bayesian Program Learning (BPL) are compared on four tasks. B) Nine human drawings (top) are shown with the ground truth parses (human) and the best model parses (machine). C) Humans and BPL were given an image of a new character (top) and asked to produce new examples. D) Humans and BPL were given a novel alphabet  and asked to produce new characters for that alphabet. E) Humans and BPL produced new characters from scratch. The grids generated by BPL are C (by row): 1, 2; D: 2, 2; E: 2, 2. Reprinted and modified from \citet{LakeScience2015}.}
\label{fig_summary}
\end{figure*}

In this paper, we review the progress made since Omniglot's release. Our review is organized through the lens of the dataset itself, since datasets have been instrumental in driving progress in AI. New larger datasets contributed to the resurgence of interest in neural networks, such as the ImageNet dataset for objection recognition that provides 1,000 classes with about 1,200 examples each \citep{Deng2009,Krizhevsky2012} and the Atari benchmark that typically provides 900 hours of experience playing each game \citep{Bellemare2015,Mnih2015}. These datasets opened important new lines of work, but they offer far more experience than human learners require. People can learn a new concept from just one or a handful of examples, and then use this concept for a range of tasks beyond recognition (Fig. \ref{fig_summary}). Similarly, people can learn a new Atari game in minutes rather than hundreds of hours, and then generalize to game variants beyond those that were trained \citep{Lake2016}. Given the wide gap between human and machine learning and the trend toward unrealistically large datasets, a new benchmark was needed to challenge machines to learn concepts more like people do.

The Omniglot challenge is to build a single model that can perform five concept learning tasks at a human level (Fig. \ref{fig_summary}). In the same paper, we introduced a framework called Bayesian Program Learning (BPL) that represents concepts as probabilistic programs and utilizes three key ingredients -- compositionality, causality, and learning to learn -- to learn programs from just one or a few examples \citep{LakeScience2015}. Programs allow concepts to be built ``compositionally'' from simpler primitives, while capturing real ``causal'' structure about how the data was formed. The model ``learns to learn'' by using experience with related concepts to accelerate the learning of new concepts, through the formation of priors over programs and by re-using sub-programs to build new concepts. Finally, probabilistic modeling handles noise and facilitates creative generalizations. BPL produces human-like behavior on all five tasks, and lesion analyses confirm that each of the three ingredients contribute to the model's success. But we did not see our work as the final word on Omniglot. We hoped that the machine learning, AI, and cognitive science communities would build on our work to develop more neurally-grounded learning models that address the Omniglot challenge. In fact, we anticipated that new models could meet the challenge by incorporating compositionality, causality, and learning to learn.

We have been pleased to see that the Omniglot dataset has been widely adopted and that the challenge has been well-received by the community. There has been genuine progress on one-shot classification, but it has been difficult to gauge since researchers have adopted different splits and training procedures that make the task easier. The other four tasks have received less attention, and critically, no new algorithm has attempted to perform all of the tasks together. Human-level understanding requires developing a single model that can do all of these tasks, acquiring conceptual representations that support fast and flexible, task-general learning. We conjectured that compositionaliy and causality are essential to this capability \citep{Lake2016} yet most new approaches aim to ``learn from scratch,'' utilizing learning to learn in ingenious new ways while incorporating compositionality and causality only to the extent that they can be learned from images. People never learn anything from scratch in this way, and thus the Omniglot challenge is not just to learn from increasingly large amounts of background training (e.g., 30 alphabets, or more with augmentation) and minimal inductive biases, only to tackle one of many tasks. Instead, the challenge is to learn from a small amount of background training (e.g., 5 alphabets) and the kinds of inductive biases people bring to the domain (whatever one conjectures those biases are), with the aim of tackling the full suite of tasks with a single algorithm. To facilitate research in this direction, we are re-releasing the Omniglot dataset with the drawing data in a new format, and we highlight two more human-like “minimal” splits containing only five alphabets for learning to learn, which we think of as more representative of human prior experience in writing and drawing. We hope our renewed challenge can drive the AI community towards more human-like forms of learning, and can encourage the cognitive science community to engage with AI in deeper ways. 

\section*{One-shot classification}
One-shot classification was evaluated in \citet{LakeScience2015} through a series of 20-way within-alphabet classification problems. Two classification trials are illustrated in Fig. \ref{fig_summary}A. A single image of a new character is presented, and the goal is to select another example of that same character from a set of 20 images produced by a typical drawer of that alphabet. Human participants are skilled one-shot classifiers, achieving an error rate of 4.5\%, although this is an upper bound since they responded quickly and were not incentivized for performance. The goal for computational models is to perform similarly or better.

Models learn to learn from a set of 964 background characters spanning 30 alphabets, including both images and drawing demonstrations for learning general domain knowledge. These characters and alphabets are not used during the subsequent evaluation problems, which provide only images. BPL performs comparably to people, achieving an error rate of 3.3\% on one-shot classification (Table \ref{table_oneshot} column 1). \citet{LakeScience2015} also trained a simple convolutional neural network (ConvNet) to perform the same task, achieving a one-shot error rate of 13.5\% by using the features learned on the background set through 964-way image classification. The most successful neural network at the time was a deep Siamese ConvNet that achieves an 8.0\% error rate after training with substantial data augmentation \citep{Koch2015}, which is still about twice as many errors as people and BPL. As both ConvNets are discriminative models, they were not applicable to the other tasks beyond classification, an ability critical to how the Omniglot challenge was formulated.

\begin{table}[t]
\footnotesize
\centering
\begin{tabularx}{3.4in}{lXXXX}
 \toprule
\multirow{2}{*}{} & 
      \multicolumn{2}{c}{\textbf{Original}} &
      \multicolumn{2}{c}{\textbf{Augmented}} \\ 
\ \ \ \ \ \ \ \ \ \ \ \ \ \ \ \ \ & Within alphabet & Within alphabet {\scriptsize (minimal)} & Within alphabet & Between alphabet
% & & & & \\
\end{tabularx}

\begin{tabularx}{3.4in}{lXXXX}
background set & & & & \\ \hline
\# alphabets & 30 & 5 & 30 & 40 \\
\# classes & 964 & 146 & 3,856 & 4,800 \\
& & & & \\
2015 results & & & & \\ \hline
Humans & $\le4.5\%$ & &  & \\ 
BPL    & \textbf{3.3\%}    & \textbf{4.2\%}    &   &    \\ 
Simple ConvNet          & 13.5\%   & 23.2\%   &   &    \\ 
Siamese Net\      &  & & 8.0\%*         &    \\ 
& & & & \\
% \medskip
2016-2018 results & & & & \\ \hline
Prototypical Net & 13.7\%   & 30.1\%   & \textbf{6.0\%}  & 4.0\% \\ 
Matching Net    &  & &   & 6.2\% \\ 
MAML   &  & &   & 4.2\% \\ 
Graph Net        &  & &   & \textbf{2.6\%} \\ 
ARC    &  & & \textbf{1.5\%*} & \textbf{2.5\%*} \\ 
RCN    & 7.3\%    & &   &    \\ 
VHE    & 18.7\%   & &   & 4.8\% \\ 
\bottomrule
\end{tabularx}
\caption{One-shot classification error rate for both within-alphabet classification \citep{LakeScience2015} and between-alphabet classification \citep{Vinyals2016}, either with the ``Original'' background set or with an ``Augmented'' set that uses more alphabets (\# alphabets) and character classes for learning to learn (\# classes). The best results for each problem formulation are bolded, and the results for ``minimal'' setting are the average of two different splits. * results used additional data augmentation beyond class expansion.}
\label{table_oneshot}
\end{table}

In the time since Omniglot was released, the machine learning community has embraced the one-shot classification challenge. Table \ref{table_oneshot} shows a summary of notable results. Among the most successful new approaches, meta-learning algorithms can train discriminative neural networks specifically for one-shot classification \citep{Vinyals2016,Snell2017,Finn2017a}. Rather than training on a single auxiliary problem (e.g. 964-way classification), meta-learning networks utilize learning to learn by training directly on many randomly generated one-shot classification problems (known as episodes) from the background set. They do not incorporate compositional or causal structure of how characters are formed, beyond what is learned implicitly through tens of thousands of episodes of character discrimination. Unfortunately it has been difficult to compare performance with the original results, since most meta-learning algorithms were evaluated on alternative variants of the classification challenge. \citet{Vinyals2016} introduced a one-shot classification task that requires discriminating characters from different Omniglot alphabets (between-alphabet classification), rather than the more challenging task of discriminating characters from within the same alphabet (within-alphabet classification; Fig. \ref{fig_summary}A). This setup also used a different split with more background characters and applied class augmentation to further increase the number of background characters four-fold, creating new classes by rotating existing classes in increments of 90 degrees. With class augmentation, the between-alphabets problem has effectively 4,800 background characters (Table \ref{table_oneshot} column 4), and meta-learning approaches have performed well (Table \ref{table_oneshot}), achieving 6.2\% using matching networks \citep{Vinyals2016}, 4.0\% using prototypical networks \citep{Snell2017}, and 4.2\% using model-agnostic meta-learning \citep[MAML;][]{Finn2017a}.

To compare our results with these recent methods, we retrained and evaluated a top-performing method, prototypical networks, on the original one-shot classification task released with Omniglot. Note that for one-shot classification, matching networks and prototypical networks are equivalent up to the choice of distance metric, and we modified the implementation from \citet{Snell2017} for within-alphabet classification.\footnote{The code's default parameters use 60-way classification for training and 20-way classification for evaluation. The default was used for the augmented within-alphabet task, but 20-way training was used for the original task since there are not enough characters within alphabets. Background alphabets with less than the required n-way classes were excluded during training. The number of training epochs was determined by the code's default early stopping train/validation procedure, except for the five alphabet case where it was trained for 200 fixed epochs.} The neural network performs with an error rate of 13.7\% (Table \ref{table_oneshot} column 1), which is substantially worse than the 4.0\% error for the between-alphabet problem. Using class augmentation to expand the number of characters within each alphabet, the network achieves 6.0\% error. Even with these additional classes, the error rate is still substantially higher than BPL, which like children can learn to learn from quite limited amounts of background experience \citep{smith-etal02}, perhaps familiarity with only one or a few alphabets with related drawing experience. BPL learns to learn so efficiently because it makes strong causal and compositional architectural assumptions, which are controversial but so far necessary for training from limited background experience (see critical commentaries from \citet{Botvinick2017a,Davis2017,Hansen2017a} and response from \citet{Lake2017BBSResponse}). To measure performance in this more human-like setting, Omniglot was released with two more challenging  ``minimal'' splits containing only five background alphabets (Table \ref{table_oneshot} column 2), and BPL still performed well (4.3\% and 4.0\% errors). In contrast, the meta-learner shows substantial degradation with minimal background training (30.8\% and 29.3\% errors), showing that meta-learning currently solves the problem in very different ways than people and BPL.

There are two other noteworthy recent architectures for constructing context-sensitive image representations. Meta-learning can be combined with graph neural networks to learn embeddings that are sensitive to the other items in the episode, achieving 2.6\% error on the between-alphabets classification task with four-fold class augmentation as before \citep[Graph Net;][]{Garcia2018}. Attentive recurrent comparators (ARCs) use a learned attention mechanism to make repeated targeted comparisons between two images \citep{Shyam2017}, achieving strong results (2.5\% error between-alphabets and 1.5\% error within-alphabets) while training with four-fold class augmentation and adding random image deformations such as scaling, shearing, translations, etc. These more complex architectures are especially at risk for overfitting, and it has been noted that training with both class augmentation and image deformations are needed for the ARC network \citep{ShyamPC}. As researchers interested in human-level learning in AI systems, we want to develop algorithms that learn with minimal training, given a rough estimate of what minimal means for people. Training with random image deformations is arguably a stand-in for invariances in the human visual system, but class augmentation is more problematic. Most people only experience one or a few alphabets through reading and writing, which is far less than the 30 provided in the \citet{LakeScience2015} background set. For the goal of reaching human-level performance with human-like training, there is a need to explore settings with both few examples per class and few background classes for learning to learn.

Another serious limitation of discriminative methods is that they only perform the task they were trained for. Human conceptual representations are far more flexible and task-general, and thus discriminative learning is not a plausible account, at least not on its own. Generative models capture more causal structure about images and can perform multiple tasks, and deep generative models have recently been applied to one-shot classification, including the neural statistician \citep[12\% error between-alphabets;][]{Edwards2016} and recursive cortical networks (RCNs), a more explicitly compositional architecture \citep[7.3\% error within-alphabets;][]{George2017}. The variational homoencoder \citep[VHE; ][]{Hewitt2018} performs well on the \citet{Vinyals2016} between-alphabet classification task with 4.8\% error, but performs much worse on the original within-alphabet classification problem (18.7\% error), which is a harder task with less background training available. Deep generative models have made important progress but they have not solved the one-shot classification problem either; they have only the barest form of causality and do not understand how real-world characters are generated, a point we discuss further in the next section.

\begin{figure*}[t]
\centering
\includegraphics[width=4.5in]{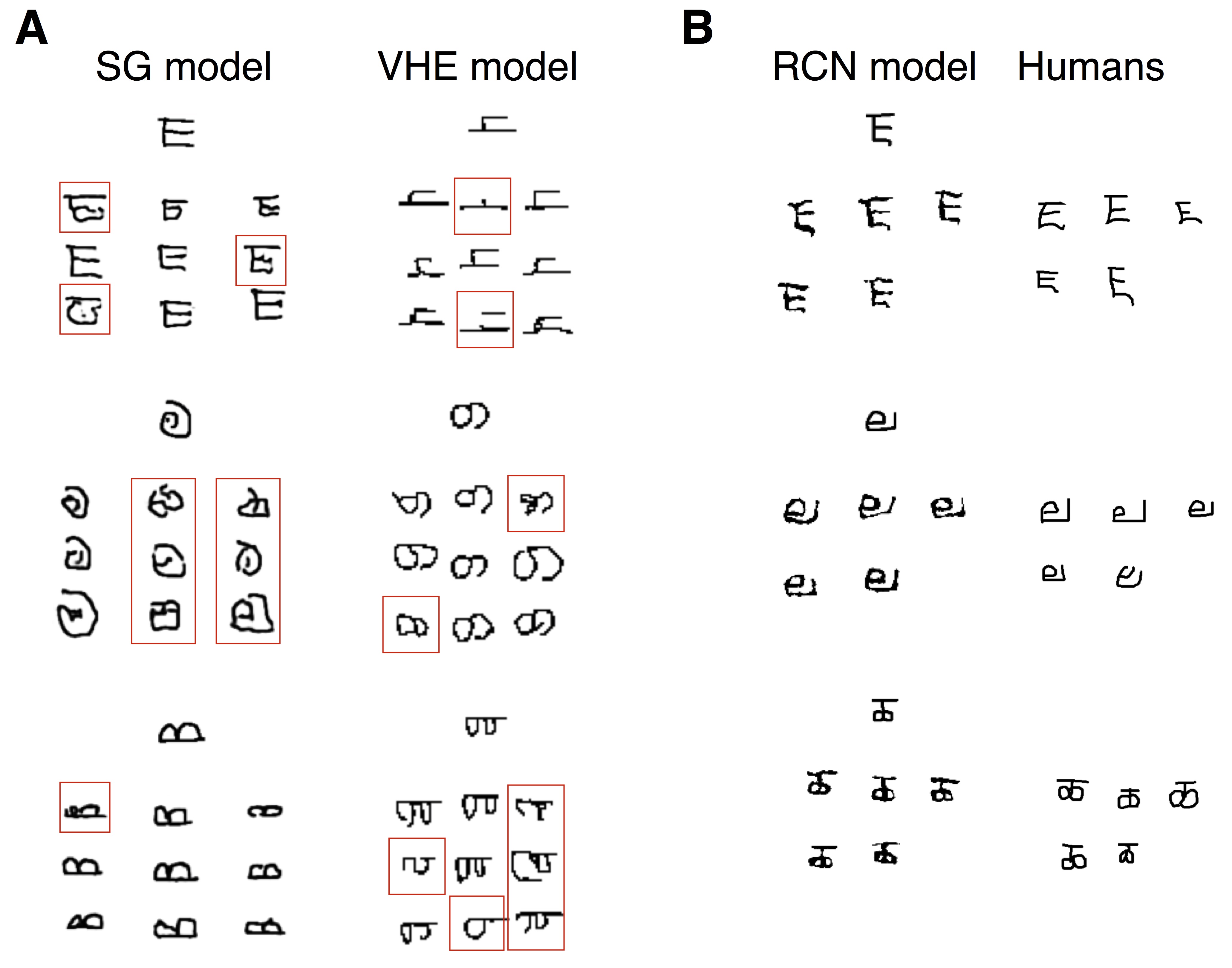}
\caption{Generating new exemplars with deep neural architectures. The task is to generate new examples (shown in grid) given an image of a new character (above each grid). A) The sequential generative model \citep[SG; ][]{Rezende2016} and variational homoencoder \citep[VHE; ][]{Hewitt2018} produce compelling examples in some cases, while showing too much variation in others (highlighted in red). B) The recursive cortical network (RCN) \citep{George2017} produces reasonable new examples but has too little variation relative to human examples from Omniglot, suggesting the model is not capturing all the degrees of freedom that people grasp in these concepts. Reprinted with permission.}
\label{fig_deepgen}
\end{figure*}

\section*{Generating new exemplars}
The Omniglot challenge is about more than classification; when a human learner acquires a new concept, the representation endows a realm of capabilities beyond mere recognition \citep{Murphy2002}. \citet{LakeScience2015} studied one-shot exemplar generation -- how people and models generate new examples given just a single example of a new concept. Human participants and computational models were compared through visual Turing tests, in which human judges attempt to determine which drawings were produced by humans and which by machines (Fig. \ref{fig_summary}C). Models were evaluated using the identification (ID) level of the judges, where ideal model performance was an ID level of 50\%. BPL can generate new examples that can pass for human, achieving an average 52\% ID level where only 3 of 48 judges were reliably above chance.

There has been substantial interest in developing generative models on the Omniglot dataset, including new neural network approaches that build on variational autoencoders, adversarial training, and reinforcement learning. Some models can generate high-quality unconditional samples from Omniglot, but it is unclear how these approaches would produce new examples of a particular concept \citep{Gregor2014,Eslami2016,Gregor2016,Ganin2018}. Other generative models have been applied to one-shot (or few-shot) learning problems, examining only generative tasks \citep{Rezende2016} or both classification and exemplar generation \citep{Edwards2016,George2017,Hewitt2018}. These approaches generate compelling new examples of a character in some cases, while in other cases they produce examples that are not especially human-like. So far, deep generative models tend to produce unarticulated strokes (Fig. \ref{fig_deepgen}A and B), samples with too much variation (Fig. \ref{fig_deepgen}A), and samples with too little variation (Fig. \ref{fig_deepgen}B). These machine-generated examples have not been quantitatively compared to human generated examples, but we are doubtful they would pass a visual Turing test.

Deep generative architectures could perform in more human-like ways by incorporating stronger forms of compositionality and causality. Current neural network models use only image data for background training, unlike BPL which learns to learn from images and drawing demonstrations. As a consequence, the networks learn to generate images in ways unrelated to how the data was actually produced, although some notable neural network models have taken more causal approaches in the past \citep{HintonNair2006}. In contrast, people have rich causal and compositional knowledge of this and many other domains in which they can rapidly learn and use new concepts \citep{Lake2012}. BPL has rich domain knowledge too and does not try to learn everything from scratch: some of these causal and compositional components are built into the architecture, while other components are learned by training on drawing demonstrations. Several recent deep generative models applied to Omniglot have taken initial steps toward incorporating causal knowledge, including using a pen or pen-like attentional window for generating characters \citep{Gregor2014,Ganin2018}. Stronger forms of compositionality and causality could be incorporated by training on the Omniglot drawing demonstrations rather than just the images. To encourage further explorations in this direction, we are re-releasing the Omniglot drawing demonstrations (trajectory data) in a more accessible format.\footnote{https://github.com/brendenlake/omniglot} The drawing demonstrations can be used in other predictive tasks, such as predicting people's motor programs for producing novel letters. BPL draws in realistic enough ways to confuse most judges in a visual Turing test of this task (Fig. \ref{fig_summary}B), although there is room for improvement since the average ID level was 59\%. We believe that building generative models with genuine causal and compositional components, whether learned or built in, is key to solving the five Omniglot tasks.

\section*{Generating new concepts}
In addition to generating new examples, the Omniglot challenge includes generating whole new concepts (Fig.  \ref{fig_summary}D and E). To examine this productive capability, human participants were shown a few characters from a novel foreign alphabet, and they were asked to quickly generate new characters that could plausibly belong to that alphabet (Fig.  \ref{fig_summary}D). BPL performs this task by placing a non-parametric prior on its programs, and judges in a visual Turing test had only a 49\% ID level in discriminating human versus machine produced letters \citep{LakeScience2015}. This ability has been explored in several deep generative architectures but with limited success, often producing blurry and unarticulated novel characters \citep{Rezende2016,Hewitt2018}. This task remains wide open challenge for deep neural networks.

The final task examines generating new concepts without constraints (Fig.  \ref{fig_summary}E). This task has received more attention and can be performed through unconditional sampling from a generative model trained on Omniglot. Many new approaches produce high-quality unconditional samples \citep{Gregor2014,Eslami2016,Gregor2016,Ganin2018}, although they have not been evaluated for their generative creativity, as opposed to merely copying characters in the training set. Nonetheless we believe this task is within reach of current neural network approaches, and the greater challenge is developing new architectures than can perform all of the tasks together.

\section*{Discussion}
There are many promising new models that have advanced the state-of-the-art in one-shot learning, yet they are still far from solving the Omniglot challenge. There has been evident progress on neural architectures for one-shot classification and one-shot exemplar generation, but these algorithms do not yet solve the most difficult versions of these problems. BPL, which incorporates more compositional and causal structure than subsequent approaches, achieves a one-shot classification error rate of 4.5\% on the original task, while the best neurally-grounded architecture achieves 7.3\% (Table \ref{table_oneshot}). On the same task with minimal background training, BPL achieves 4.2\% while the best neural network results are 23.2\% errors. The more creative tasks can be evaluated with visual Turing tests \citep{LakeScience2015}, where ideal model performance is a 50\% ID level based on human judges. BPL achieves an ID level of 52\% on one-shot exemplar generation (and 55\% with minimal background training), 59\% on parsing new examples, 49\% on generating new concepts from a type, and 51\% on generating new concepts without constraints. The Omniglot challenge is to achieve similar success with a single model across all of these tasks jointly. 

Some of the most exciting advances in the last three years have come from using learning to learn in innovative ways. A similar sustained and creative focus on compositionality and causality will lead to substantial further advances. We have yet to see deep learning approaches that can achieve human-level performance without explicitly making use of this structure, and we hope researchers will take up the challenge of incorporating compositionality and causality into more neurally-grounded architectures. This is a promising avenue for addressing the Omniglot challenge and for building more domain-general and more powerful human-like learning algorithms \citep{Lake2016}.

Human concept learning is characterized by both broad and deep expertise: broad in that people learn a wide range of different types of concepts -- letters, living things, artifacts, abstract concepts, etc., and deep in that people learn rich representations for individual concepts that can be used in many different ways -- for action, imagination, and explanation (Fig. \ref{fig_summary}). Both forms of expertise are essential to human-level intelligence, yet new algorithms usually aim for breadth at the expense of depth, targeting a narrow task and measuring performance across many datasets. As a representative case, matching networks were applied to three different datasets but only to one task, which was one-shot classification \citep{Vinyals2016}. Deep conceptual representations remain elusive in AI, even for simple concepts such as handwritten characters; indeed \citet{Hofstadter1985} famously argued that learning to recognize the characters in all the ways that people do contains most of the fundamental challenges of AI. As \citet{Mitchell2018} put it recently, ``artificial intelligence [has hit] the barrier of meaning,'' in that machines do not understand the world as people do.

Cognitive scientists can help break this barrier. They have expertise in studying how much people know about a domain, such as handwriting or drawing, and how richly they know it. Cognitive scientists can lead expeditions into new domains, which like Omniglot could be introduced as new challenges for AI. As with the Omniglot challenge, the goal would not be to tackle just one task across many domains, or to tackle separate tasks with separate models. Achieving human-level learning requires tackling a range of tasks together by learning deeper, more flexible, and more human-like conceptual representations, such that one model can seamlessly perform many different tasks.

Human-level understanding includes the five tasks discussed here and many more. These five representative tasks are surely an important start, yet more tasks and benchmarks would further accelerate progress -- another promising avenue for cognitive scientists and AI researchers to pursue together. Several novel and interesting classification tasks with Omniglot have already been contributed: \citet{Santoro2016} and \citet{Rae2016} studied sequential one-shot classification where stimuli arrive sequentially, and \citet{Woodward2016} studied an active learning version of the same task. Other more challenging versions of within-alphabet classification should be studied too. We especially encourage new Omniglot tasks that go beyond classification, and new directions could include filling in occluded images, understanding CAPTCHAs constructed with novel characters, classifying new characters by alphabet, or constructing new characters from verbal descriptions. Each new task offers an additional bridge between cognitive science and AI, with potential for human behavior and cognitive principles to inform the development of new algorithms. We are excited to see what additional progress the next three years will bring.

\subsection*{Acknowledgements}
We are grateful to Jason Gross for his essential contributions to Omniglot, and we thank the Omniglot.com encyclopedia of writing systems for helping to make this dataset possible. We thank Kelsey Allen, Reuben Feinman, and Tammy Kwan for valuable feedback on earlier drafts, and Pranav Shyam for his helpful correspondence regarding the ARC model.

\bibliographystyle{apa}
\interlinepenalty=10000
\bibliography{library_clean,pc}
% \bibliography{annotated,library_clean,pc}
\end{document}